\documentclass[letterpaper]{article} % DO NOT CHANGE THIS
\usepackage{aaai23}  % CHANGE THIS BACK TO LINE ABOVE BEFORE SUBMITTING!!
\usepackage{times}  % DO NOT CHANGE THIS
\usepackage{helvet}  % DO NOT CHANGE THIS
\usepackage{courier}  % DO NOT CHANGE THIS
\usepackage[hyphens]{url}  % DO NOT CHANGE THIS
\usepackage{graphicx} % DO NOT CHANGE THIS
\usepackage{natbib}  % DO NOT CHANGE THIS AND DO NOT ADD ANY OPTIONS TO IT
\usepackage{caption} % DO NOT CHANGE THIS AND DO NOT ADD ANY OPTIONS TO IT
\usepackage{algorithm}
\usepackage{algorithmic}

\usepackage{newfloat}
\usepackage{listings}
\DeclareCaptionStyle{ruled}{labelfont=normalfont,labelsep=colon,strut=off} % DO NOT CHANGE THIS
\floatstyle{ruled}
\newfloat{listing}{tb}{lst}{}
\floatname{listing}{Listing}
\usepackage{textcomp}

\usepackage[utf8]{inputenc}

\defcitealias{kite:1}{Kite}
\defcitealias{copilot:1}{GitHub Copilot}
\defcitealias{copilot:2}{GitHub Copilot FAQ}
\defcitealias{codex:1}{OpenAI Codex}

\title{Autonomous Intelligent Software Development}
\author {
    Mark Alan Matties
}
\affiliations{
  The Johns Hopkins Applied Physics Lab \\
  11100 Johns Hopkins Rd. \\
  Laurel, MD 20723 \\
  mark.matties@jhuapl.edu \\
}

\begin{document}

\maketitle

\begin{abstract}
We present an overview of the design and first proof-of-concept implementation for AIDA, an autonomous intelligent developer agent that develops software from scratch. AIDA takes a software requirements specification and uses reasoning over a semantic knowledge graph to interpret the requirements, then designs and writes software to satisfy them. AIDA uses both declarative and procedural knowledge in the core domains of data, algorithms, and code, plus some general knowledge. The reasoning codebase uses this knowledge to identify needed components, then designs and builds the necessary information structures around them that become the software. These structures, the motivating requirements, and the resulting source code itself are all new knowledge that are added to the knowledge graph, becoming available for future reasoning. In this way, AIDA also learns as she writes code and becomes more efficient when writing subsequent code.
\end{abstract}

\section{Introduction}
Software development as practiced today is still a largely manual activity that is time consuming, costly, and error prone, depending on the skill of the human software developers (``developers''). All good, competent software development requires a high level of skill, but much of the ``day to day'', routine programming is not necessarily creative. It is done by employing well used patterns across a range of detail levels, from the very high-level of overall program design to very low-level of writing syntactically correct statements. Oftentimes, the only meaningful variation from one program to the next is some context, such as the variable names chosen based on the type of data they hold. If these patterns can be encoded \textit{and} the contextual variations discerned and addressed, we claim that much of routine software development can be \textit{fully automated}, taking the human completely out of the writing code loop.

These characteristics make software development ripe for intelligent automation. Limited progress has already been made in the forms of (1) code templating, (2) ``intelligent'' completion, and (3) some early forays into machine learning (ML) systems that write code. Code templating is usually targeted, and even wide-scale code reuse remains a largely unfulfilled promise. Intelligent completion systems like Kite \citepalias{kite:1} and Copilot \citepalias{copilot:1} use ML models to reduce the amount of typing a developer must do by guessing what she intended to type and filling it in, reducing the amount of typing she must actually do. However, neither removes her from the loop. Rather, they seem to turn her into a real-time reviewer/debugger of the ``completer’s'' code. A ``Frequently asked questions'' section of the Copilot website states that ``In a recent evaluation, we found that users accepted on average 26\% of all completions shown by GitHub Copilot.'' \citepalias{copilot:2} So, in practice, the benefit is questionable. 
	
Platforms that use ML models to write code, such as OpenAI Codex \citepalias{codex:1}, are still bound by the inherent limitations of ML. They include the requirement of a very large training data set of (existing) code samples, some of which may be of questionable quality, and that ML methods are generally much better at interpolating than extrapolating. In this context, interpolating translates to copying ``known'' (ie, learned through training) code structures with limited variation rather than creating new code structures, which would translate to extrapolating. Importantly, they do not easily permit the introduction of new program structures or programming patterns, and certainly without a large library of existing samples and retraining of the model.

The first two approaches (templating and completion) still require a great deal of typing by developers to customize parts of what automatically gets filled in. Picking variable names or stringing together function calls of the implemented algorithms are not exercises in creativity, as they also follow patterns that draw from general knowledge. The third approach is still early in its exploration, but has some underlying limitations.

These constraints and deficiencies provide the motivation for AIDA, the \textit{autonomous intelligent developer agent}. The overarching goal of AIDA (``$aye$-$ee$-$duh$'') is to produce software at the level of an expert human developer, given only a problem statement or set of requirements. AIDA reasons over encoded knowledge to build information structures, which in turn become part of her available knowledge. Much of the knowledge is specific and deals in patterns of various kinds across multiple scales of detail. Some of it is simply general and allows AIDA to find and use needed components or add customized stylistic qualities. 

AIDA writes code from scratch using semantic reasoning in a top-down approach. She reasons from general to more specific terms about how to develop software, rather than blindly following patterns gleaned from a multitude of code samples. This approach enables AIDA to not only write correct, high-quality code, but also to easily incorporate style elements using general knowledge. Ultimately, AIDA's purpose is to empower anyone who needs software but is not a programmer to create high quality software very quickly and at a very low cost simply by providing proper requirements. 

In this paper, we present an overview of the design and first working version of AIDA. This version is intended to explore whether our approach is at all feasible, and so the knowledge currently encoded, especially the programming patterns, are necessarily limited. We believe that our approach is generalizable and extendable to fuller, much richer representations that will lead to creating substantial analytic software that is useful for real world applications.

\section{Background}
\subsection{Semantic Reasoning}
In this paper, we view semantic reasoning broadly as the processing of linguistically meaningful information using methods drawn from the field of logic. We are especially interested in the now popular \textit{knowledge graph} \cite{singhal2012introducing,ehrlinger2016towards} approach, which views information in terms of entities and the relationships between them. Two entities linked by a relationship forms a statement to which one can assign a truth value. By virtue of being in the knowledge graph, the statement has a truth value of ``\textit{True}''. This view naturally lends itself to a graphical representation, where the entities are nodes and their relationships are edges, hence the knowledge graph paradigm. Currently, there are a few different forms that a knowledge graph can take in practice. We use the one where a knowledge graph is built up from OWL2 \cite{OWL2Profiles} description logic \cite{abs-1201-4089} ontologies, which, in this work, we construct from expert human knowledge.

We import the ontologies into the knowledge graph, where they serve as a sort of schema. The knowledge graph is progressively built out through the addition of of knowledge (``\textit{instance}'' data) according to the structure defined in the ontologies. In addition to the graph itself, a semantic reasoning platform requires a \textit{reasoning codebase}, which reads information from the knowledge graph, performs the logic operations, eg, inference, and (possibly) inserts new information back into the knowledge graph. For AIDA, the knowledge graph and specific reasoning codebase are inextricably interwoven.

\subsection{Encoding Knowledge for Semantic Reasoning}
As currently practiced, knowledge graphs encode only \textit{declarative knowledge} (``\textit{knowing-that}''), which one can view as ``simple facts''. For example, we can create two logical statements (in \textit{subject predicate object} format)
\begin{verbatim}
  Walter_Pitts   born-in   Detroit
  Detroit        city-in   Michigan
\end{verbatim}
and assign the value true to each of them. In practice, any information encoded in such a knowledge graph is taken to be true by virtue of its inclusion and does not change unless it is later shown, in fact, to be false. Hopefully, such cases are rare. Then, a knowledge graph is mostly static unless new facts are added to it.

In addition to declarative knowledge, epistemology entails the study of \textit{procedural knowledge} (``\textit{knowing-how}'') \citep{sep-knowledge-how}. Encoding procedural knowledge is challenging because, as we learn from introductory logic courses, we may assign a truth value only to declarative statements, and so use them in logical reasoning. We normally think of procedural knowledge in terms of imperative statements, eg, ``drive three miles, make a right turn, drive two more miles, look to your left''. Unfortunately, imperative statements cannot have a truth value and so cannot be used in logic, at least in that form. However, ``knowledge'' about ``how to do things'' is implicitly encoded in the reasoning codebase that works with the knowledge graph proper. If we can recast procedural knowledge, as least in part, into declarative statements, we can move that part of it into the knowledge graph. Through this addition of procedural knowledge to semantic reasoning, AIDA can build information structures, not just infer and report back simple facts. As far as we know, this approach to explicitly encode some part of procedural knowledge in the ontologies is novel.

\section{Design and Architecture}
\subsection{Design Considerations}
First and foremost, AIDA must write code that is both technically and contextually correct. By technically correct, we mean that the resulting code must run without error and produce correct results as stated in the requirements. It must be \textit{verifiable} by current conventional means. We expect that until AIDA gains some level of trust from users, any software she writes would be subjected to code reviews by people. So, it must be readable by people without undue cognitive strain. 

By contextually correct, we mean that stylistic elements should conform to accepted standards and/or stated preferences, such as would be found in a style guide. For example, labels such as variable names should be chosen that are meaningful in context while also conforming to programming language constraints, such as a maximum character length. A variable in a Python program holding the value of a temperature reading should be named ``\texttt{current\_temperature}'' not ``\texttt{T}''. Other contextual elements include (1) customizable ordering, such as the order of listing imported code libraries (eg, alphabetical by library name), (2) use of name aliases, and (3) even preferred patterns or structures, eg, using \textit{list comprehension} in Python. 

\begin{figure}[t]
\centering
\includegraphics[width=0.99\columnwidth]{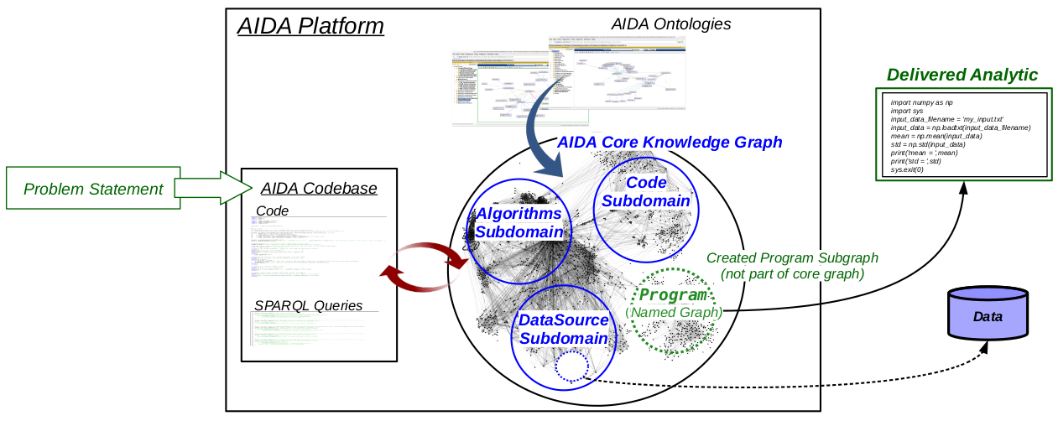}
\caption{A high-level view of the AIDA architecture.}
\label{fig_aida_architecture}
\end{figure}

Finally, the code must be easy to maintain, so documentation is a key element. Any documentation must be an integrated part of the software AIDA develops. Using the knowledge graph, the documentation can be co-located with the requirements, any and all built structures, and the source code itself. Documentation then need not be slapped together as an afterthought, but can be built and modified in real time as the software is built. As \textit{natural language generation} improves and matures, we expect that AIDA can even interface with it to write user guides by providing context-aware information to it.

\subsection{Architecture Considerations}
Since AIDA must build software from knowledge about how to build software, we took a semantic reasoning design approach that has been extended to use both declarative and procedural knowledge, as shown in Figure \ref{fig_aida_architecture}. Both core and general knowledge are explicitly encoded in a set of ontologies that are imported into and form the basis of the knowledge graph. How and when to use that knowledge is implicitly encoded in the reasoning codebase. It consists of a set of SPARQL queries to read from and write to the knowledge graph, as well as the code to perform these actions and manipulate the results. Both components - the knowledge graph and the reasoning codebase - form the extended semantic reasoning platform that is AIDA.

The three ``pillars'' of AIDA's core knowledge are \textit{data}, \textit{algorithms}, and \textit{code}. Her core knowledge is in those domains that are specifically tied to writing software. In our view, nearly all software exists to take some input, apply one or more algorithms to it, and transform the input to some output. In the current version of AIDA, the input comes from some external data source, such as a data file. Before starting to write code, AIDA must have detailed, specific knowledge about the needed data - what it describes, how and where to get it, its format, the nature of it content, etc. Likewise, she must have specific information about algorithms - what kind of transformation they perform, what are their input and output types, what are their time and/or space complexities, etc. Note that AIDA does not store any raw data, as shown in Figure \ref{fig_aida_architecture}. Doing so would quickly overwhelm her knowledge base, making it unusable. Rather, she holds all the detailed \textit{metadata} that one would need to use the data when writing software and the data resides wherever it resides external to her, including a description of (ie, pointer to) that location. 

For algorithms, AIDA currently encodes knowledge \textit{about} how and when to use them, not any detail about their constituent operations. This knowledge includes the type of algorithm across several different aspects, the purpose of it, ie, ``type of calculation'' it performs, the required inputs, and the output it produces. Other knowledge, such as algorithmic complexity, is included as well. Again, all the metadata a developer would need to choose an algorithm, AIDA requires that knowledge as well. We only briefly describe knowledge about algorithms here, deferring the details to a future manuscript. We can include the more detailed knowledge in the future, when it becomes useful. Then, she should be able to mix and match individual operations and develop variations on algorithms, if not entirely new ones.

Of course, AIDA must know about code – not only about low-level, detailed programming language structures and when to use them, but also about high-level, programming language agnostic information, such as gross program structure. In fact, all programming language structures are implementations of abstracted ones that are language agnostic. AIDA exploits this relationship and deals in both.

As shown in Figure \ref{fig_aida_architecture}, AIDA takes in a problem statement and analyzes it to determine what program to write. The problem statement must specify some minimal information about the desired program, especially what output is needed from it. A user might also state what input data or algorithms to use in either specific or general terms. For example, ``use the data set with name \texttt{my\_input.txt} and calculate the average value of the data with an error estimate''. The "data set" part of the specification is quite specific. The "calculate" part of it actually states both the desired output and (indirectly) the type of algorithm to use. It is somewhat more general and descriptive of the kind of algorithm to use (such as \textit{average value}), rather than naming a specific one (such as \textit{arithmetic mean} or \textit{geometric mean}). Likewise, the user may state desired aspects of the desired software itself. Usually, the user should at least specify the programming paradigm, such as imperative/procedural, and one or more programming languages into which AIDA should render the program, such as Python or Python-3.8.10. The user might also specify a preference for a certain external code library, such as NumPy. These are just a few simple examples of many possible variations.

If AIDA finds multiple matches for any of these specifications, she must have some criteria to choose among them. The criteria could also be specified by the user, or AIDA could know about default selection criteria. While we try to keep code details separated from data and algorithms, some dependency is inevitable for highest efficiency. For example, selection criteria might be to choose the data whose source is compatible with, ie, can be read by, a given code function. Or, the user might specify that the least expensive data source be used, if monetary cost of the data were an attribute known to AIDA. For algorithms, a common criterion is the time complexity of the algorithm, though other efficiency metrics are possible.

\subsection{Architecture of Information Structures}
All programming languages (PLs) provide a means to render a set of abstract information structures into runnable code. Which structures to use and how to arrange them is driven by both hard and soft constraints. Oftentimes, a given PL is designed for a certain programming paradigm, such as procedural or object oriented. The choice of a particular paradigm drives much of the overall structure of the software and, ultimately, the PL choice. Regardless, the PL will implement all abstract structures necessary to render source code in that particular paradigm (and if it cannot, AIDA can have that knowledge as well).

We designed AIDA to have knowledge about both abstract structures and programming language implementations of them. For maximum flexibility, AIDA builds software in two stages. After parsing the requirements, she first builds a representation using abstract structures that depends on the programming paradigm, but is programming language agnostic (the \textit{PLA version}). From the PLA version, she builds a programming language rendered (PLR) version in the chosen programming language. Using this two stage approach, AIDA is designed to correctly render the same PLA version into multiple programming languages without the problems of ``translating'' between different PLs (eg, Java / C++) or even versions of the same PL (eg, Python 2.x / Python 3.x). In our approach, the abstract structures in the PLA version of the software are faithfully mapped to any supported PL, so that it can be rendered ``naturally'' in that language. Any quirks or special requirements of a PL can be part of AIDA's knowledge. Both the PLA version and any PLR version that AIDA constructs become part of her knowledge and can be used in future projects. Finally, AIDA ``walks'' the PLR version to write the source code out to a file. 

Authoring software requires knowledge about much more than software structures and programming language syntax. Distilled to its essence, software is created to take information you ``have but don't really need'' and transform it into information ``you really need but don't have''. For example, software that predicts the closing price of some stock takes historical price data over the course of the day (information you have really but don’t need) and transforms it into a closing price for the day (information you really need but don’t have). It performs this transformation using algorithms that have been implemented in code.

\begin{figure}[t]
\centering
\includegraphics[width=0.99\columnwidth]{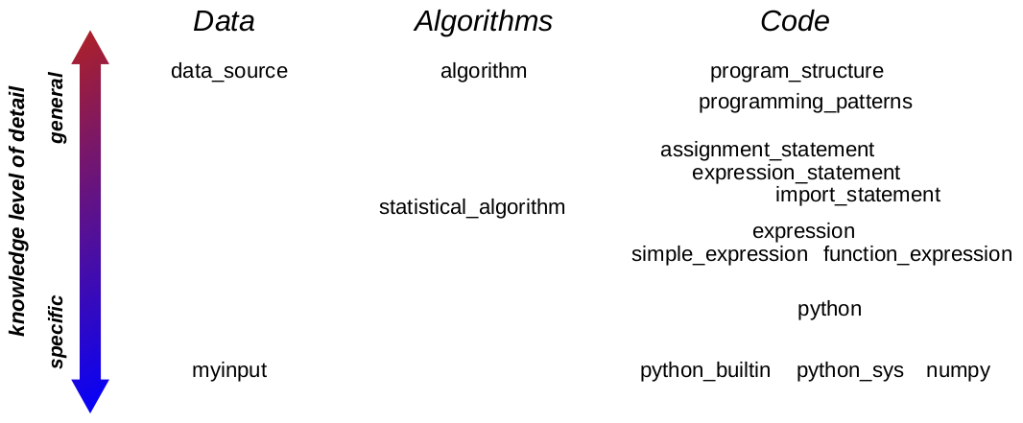}
\caption{A high-level view of the AIDA ontologies.}
\label{fig_aida_ontologies}
\end{figure}

\section{Implementation}
In practice, most of what AIDA does is (1) find and match attributes of different kinds of entities through relationships, and (2) create and modify patterns of structures. Knowledge about the entities and their relationships is encoded in the knowledge graph via OWL2 ontologies. We developed all ontologies from scratch using Protege \cite{protege-Musen15}. While there are many ontologies published across a wide range of knowledge domains, we have not found any that could substantially be used for AIDA. In our opinion, the content of an ontology, especially its object properties, is driven by its intended application. It seems to us that the intended application of many published ontologies is to demonstrate that their authors are experts in the hierarchy of some field, as they are too often little more than taxonomies. We believe the real power of an ontology comes from its ``cross-branch'' relationships (expressed as object and datatype properties), not its taxonomic hierarchy.

The criteria of which attributes of which entities should be matched for which purposes is currently held in the logic of the reasoning codebase. As described above, AIDA's procedural knowledge is currently spread across both the KG and the RC. The part of the procedural knowledge that is about the structures of the processes, such as the names of their constituent steps and, especially, their order of execution are held in the knowledge graph. The statements that execute the constituent steps are still encoded in the reasoning codebase. We are actively working on how to cast all procedural knowledge so that it can be encoded to the greatest degree possible in the KG.

To encode the necessary knowledge, we took the design goals described above and created a set of OWL2 ontologies covering the domains of data, algorithms, and code, as well as some supporting ontologies containing general knowledge. For our manageability, we break the ontologies up into smaller units (ie, files) that are cross-linked through inclusion of others' entities and relationships. All ontology files are imported into AIDA's graph database to form her initial knowledge graph. Figure \ref{fig_aida_ontologies} shows the core ontologies we created for the first prototype implementation. (There are several other supporting ontologies that provide general knowledge about what is a quantity, a value, units of measure, and more.) The \textit{Data} ontologies cover the range of knowledge needed to find and use specific data in practice. At the highest level, the \textit{data\_source} ontology holds knowledge about all possible classes of data sources, especially the \textit{content} of data held and \textit{container} in which they are held. Here, there is only one specific data source. It is described in the ontology \textit{myinput}. It describes an ASCII encoded file of six floating point numbers arranged in CSV format with zero header rows and six rows of data with the filename \texttt{my\_input.txt}.

The ontologies under \textit{Algorithms} are also few in number. In this first implementation, the \texttt{algorithm} ontology primarily encodes high-level information, such as that an algorithm requires some inputs (number and types), that it produces output (number and types), a description of what kind of transformation it performs, and what its ``big-O'' time complexity is. For example, the two algorithms used in the first implementation are the \textit{arithmetic mean} and the \textit{standard deviation}. In order to choose these algorithms, AIDA must know at a minimum that they require input that is numerical, consists of at least two values, and that all values must of the same quantity type. That is, it makes no sense to calculate the arithmetic mean of a temperature value, a humidity value, and a pressure value, which are different types of quantities. The algorithm produces an output that is a single value that is of the same quantity type as the input, and is an average value of the input values. At this stage, the label ``average value'' is just that, a label. AIDA does not really ``know'' what it means except that it is a kind of output value of an algorithm and also a string found in the parsed problem statement, and that they match in this context. However, as AIDA gains more instances of using the ``average value'' label (ie, matches that label in different contexts), she better ``understands'' its scope, which means she can use that knowledge across a wider range of logical arguments.

In the discussion above, we stated our goal of AIDA reasoning over both declarative and procedural knowledge. While not commonly viewed in this way, declarative knowledge is normally explicitly encoded in the knowledge graph and procedural knowledge, such as it is, is implicitly encoded in the associated reasoning codebase. It is not normally viewed in this way because, as in the case of publicly available general knowledge graphs, the reasoning codebase is left up to the consumer of the knowledge graph. 

There is not usually such a tight integration and interdependence of the knowledge graph and the reasoning codebase. In this version of AIDA, the procedural knowledge is not yet completely located within the knowledge graph. It is split across the knowledge graph and the reasoning codebase. The ``structural`` part of the procedural knowledge consisting of certain structures and their elements and, especially, their dependencies and ordering, are recast as declarative knowledge and encoded in the knowledge graph (through ontologies). The actual ``doing'' is still implicit in the reasoning codebase, along with some knowledge about the when to do the ``doing''. 

We illustrate the division of knowledge with a simple example -- building the Python statement \texttt{import sys}. We begin at the point where AIDA has already identified that she used the \texttt{sys} package elsewhere in the program and now needs to build the statement to \texttt{import} it. In order to build the statement, AIDA needs to know (1) which variation of the statement she should use, (2) what are the elements of the one she selects, and (3) what is the order of the elements in the statement. AIDA uses this knowledge to build the statement and insert it into the correct part of the PLR representation of the program. All the information in (1) -- (3) is encoded in the knowledge graph (KG). The knowledge that the order is a given is implicit in the reasoning codebase (RC), because the information is queried in that order in the RC. 

While we defer the full details to a later paper, the process AIDA follows is to execute the correct query to get the information in  (1) -- (3), where the details of the next query is dependent on the results of the previous one.  Once AIDA retrieves all this information, it becomes input to the part of the RC that builds the statement. She then executes a query to find the correct place in the PLR structure and executes a final query (for this process) to insert the new information in the KG. 

This high-level example illustrates the tight connection between AIDA's KG and RC. It suggests where we can continue to move procedural knowledge from the RC to the KG. For one, we can encode the order of (1) -- (3) above, even encoding the entire query for each step in the KG. As we have more examples of what AIDA must do, ie, what procedural knowledge she uses to what ends, we can encode those patterns into the KG in more detail. Then, AIDA could build her queries dynamically as needed and retain the ones she uses most often in the KG for efficiency. We are actively working on creating this capability.

\begin{figure}[t]
\centering
\includegraphics[width=0.99\columnwidth]{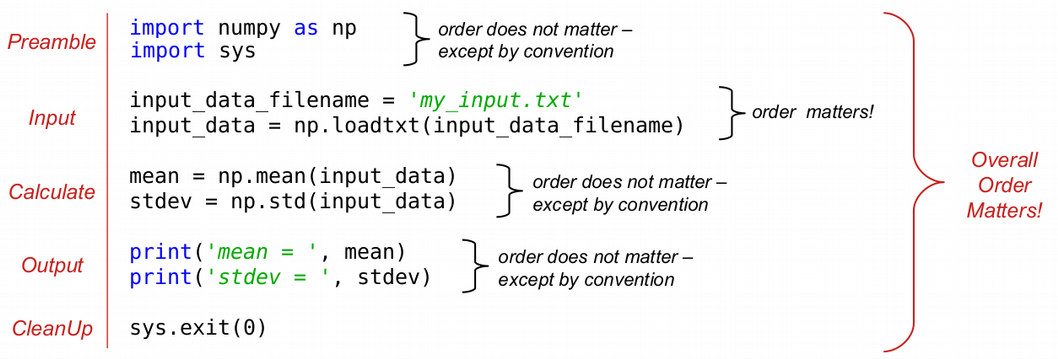}
\caption{The annotated target program decomposed into program sections. The section names appear in red to the left and are the components of the program structure type \texttt{Input\_Calculate\_Output}.}
\label{fig_aida_target_program}
\end{figure}

\section{First Exemplar Problem and Results}
Obviously, the problem area we are attacking is very wide ranging and a first implementation of it using the approach we outline above can become very large, complicated, and overwhelming very quickly. For the first implementation of AIDA, we created a very simple exemplar problem with a target program, show in in Figure \ref{fig_aida_target_program}, that AIDA should write given the problem statement in Figure \ref{fig_aida_problem_statement}. This program contains what we consider to be a set of bare minimum elements constituting analytic software. 

The core elements are that it must (1) read some data, (2) apply an algorithmic transformation, and (3) produce some report of the results. The ancillary, but still necessary, elements are that it must import external code libraries and explicitly exit the program. It should also exhibit some minimal examples of customization. We used the target program as a general guide in scoping our ontologies. in developing the content of the ontologies, we took an approach of ``reverse engineering'' the target program with a wide view. Essentially, we decomposed the actions across all levels one would take to conceive of and write the target program into general patterns, as one might find in a textbook on programming.

\begin{figure}[t]
\begin{lstlisting}[numbers=none,upquote=true,xleftmargin=0cm]
data_sources_names = ['my_input.txt']
requested_calculations = \
       ['average value', 
        'average value variation']
program_requirements = \
      ['read input data', 
       'calculate quantity', 
       'report result']
programming_language = 'Python-3.8'
program_basename = 'hello_analytic'
\end{lstlisting}
\caption{The first target program for AIDA to compose.}
\label{fig_aida_problem_statement}
\end{figure}

The target program performs the minimal functions required as described. It has additional stylistic and contextual components, such as ordering \texttt{import} statements alphabetically by ``official'' package name (ie, not alias), use of a package name alias (\texttt{np} for \texttt{numpy}), and use of meaningful variables names. We emphasize that the variable names are not ``hard coded''. They come from general \textit{patterns} encoded in AIDA's knowledge graph about how to pick variable names. In this first simple example, there are three patterns. The first is that a literal is assigned to a variable and that literal is the name of a data source file, then the name should be a combination of its types. In this case, those types are ``input\_data'' and ``filename'' for \texttt{input\_data\_filename}. The second is that if a ``data source filename'' variable is used as an argument to a function that reads its data, the variable name should be just the type of data. Here, it is ``\texttt{input\_data}''. Lastly, if the return value of a function is assigned to a variable, the name of the variable should be the same as the name of the function, such as ``\texttt{mean}'' and ``\texttt{std}''. Since the program is an example of a prototypical analytic and is written in Python, we named it \texttt{hello\_analytic.py}. Note that AIDA knows the fact that a Python program should have the extension ``\texttt{.py}'' and appends it to the base program name when she writes out the source code to a file. Importing the ontologies and building the starting knowledge graph takes about 75 seconds, since this knowledge graph is small.

In the problem statement, we identify a specific data source to use (in \texttt{data\_source\_names}), name the type of calculations to perform (in \texttt{requested\_calculations}), and give the program requirements, ie, what the program should do (in \texttt{program\_requirements}). We also specify the programming language (in \texttt{programming\_language}). Note that the problem statement is quite simple. Each of the strings in the problem statement appear in the knowledge graph as labels to some entity (with the exception of \texttt{program\_basename}) in order to prove out the simplest matching. While certainly not unimportant, the richness of the problem statement is not a focus at this stage of our work. It will grow as AIDA's capabilities grow.

From this statement, AIDA should find that \texttt{myinput} holds the detailed metadata of a specific data source (as opposed to a type of data) with the name ``\texttt{my\_input.txt}''. AIDA gathers all the metadata about this data source. Likewise, AIDA should (and does) find that each of the two \texttt{requested\_calculations} are descriptions of the types of calculations and that they map to the specific algorithms \texttt{arithmetic\_mean} and \texttt{standard\_deviation} (the only such algorithms that she knows). AIDA checks that the requirements on the inputs to these algorithms (such as that there are two or more values and that all the values have the same quantity type) match the data source she found. In fact, they match (by design). Finally, AIDA will find from the `\texttt{report result}` part of \texttt{program\_requirements} that she should ``print out'' the results. 

Having found entities for the data source and algorithms, AIDA reasons over her knowledge to build the code structures from high level to low level. At the highest level is the program structure that satisfies the stated \texttt{program\_requirements}. It is \texttt{Input\_Calculate\_Output}, which is a subclass of \textit{ProgramStructure}. At the intermediate level are the code functions that can read the data source and implement the identified algorithms. Respectively, they are \texttt{loadtxt}, \texttt{mean}, and \texttt{std}, all from the \texttt{numpy} Python package. At the lowest, most detailed level is building the needed lines of code (ie, statements) with the correct structure and and writing them out in the correct order. All of these entities have detailed metadata encoded in AIDA's constituent ontologies, such as would be needed for a developer to use them.

Note that in the KG, we can specify different ordering for different purposes. The sections of the chosen program structure are \texttt{Preamble}, \texttt{Input}, \texttt{Calculate}, \texttt{Output}, and \texttt{CleanUp}. When AIDA writes out the source code to a file, it must have this order. The program must \texttt{import} packages before using them, etc. However, when \textit{composing} the program, it is better to do so in the order, \texttt{Input}, \texttt{Calculate}, \texttt{Output}, \texttt{CleanUp}, \texttt{Preamble}. Then, dependencies from one section can flow naturally into the next. For example, the \texttt{Input} section names the variable that holds the data (input\_data) that is needed in the \texttt{Calculate} section. However, all sections after \texttt{Preamble} might have functions from external code libraries. While those sections are composed, AIDA makes note of the external code libraries that are referenced and then creates the \texttt{import} statements when all the other sections are complete.

In this first simple step for AIDA, each of these calculations has only one matching algorithm, namely \texttt{arithmetic\_mean} for ``\texttt{average value}'' and \texttt{standard\_deviation} for ``\texttt{average value variation}''. In the future ,when more knowledge is encoded in the knowledge graph, AIDA will need methods to make choices, not only on algorithms, but also on data sources, code functions and other structures. Such methods are essentially algorithms themselves and so could be encoded in the knowledge graph.

\begin{figure}[t]
\begin{lstlisting}[numbers=none,upquote=true,xleftmargin=0cm]
import numpy as np
import sys
input_data_filename = 'my_input.txt'
input_data = np.loadtxt(input_data_filename)
mean = np.mean(input_data)
std = np.std(input_data)
print('mean = ',mean)
print('std = ',std)
sys.exit(0)
\end{lstlisting}
\caption{The literal, unaltered source code written out by AIDA to a file named \texttt{hello\_analytic.py}.}
\label{fig_aida_composed_code}
\end{figure}

The actual source code that AIDA writes to satisfy this problem statement is given in Figure \ref{fig_aida_composed_code}. It is not reformatted in any way. It matches the target program almost exactly, except for very minor stylistic elements, such as an empty new line between program sections and a space between the comma and the variable name in the \texttt{print} statements. (These elements can also be included, they just have not been yet.) This source code is correct and gives the correct output when run (with the correct Python interpreter). We give some statistics on the AIDA v0.1 implementation in Figure \ref{fig_aida_statistics}. The core graph is built from 20 ontologies and its core size, ie, before AIDA builds any code, is more than 22,500 triples. The two versions of the program (PLA and PLR) combined add up to almost 1500 triples in the graph, split more or less evenly. They are stored in a \textit{named graph} subgraph for easy reference and access. On a modest laptop, AIDA takes about 5 minutes to build the knowledge graph from scratch, build the PLA and PLR versions of the program, and write out the source code to a file. 

\begin{figure}[t]
\begin{small}
\begin{verbatim}
  Ontologies – 20 ontologies
  Knowledge Graph
     Total “core” size > 22,500 triples
     Named graph size = 1458 triples
     PLA version = 677 triples
     PLR version = 546 triples
  Reasoning Codebase
     Lines of Python code > 4300
     Lines of SPARQL queries > 1900
     Number of SPARQL queries = 129
     Number of SPARQL query calls = 107
Total time to run ~ 5 minutes on laptop
\end{verbatim}\end{small}
\caption{Statistics on the AIDA v0.1 implementation.}
\label{fig_aida_statistics}
\end{figure}

Thus, we have successfully demonstrated our concept for and implementation of an extended semantic reasoning platform to develop software from scratch for a minimal exemplar program.

\section{Future Work}
We have already extended AIDA's capability to compose a composite algorithm from several simple ones (manuscript in preparation). Currently, we are working to give AIDA the ability to use other basic software structures of loops and conditionals. The challenge is not in the building of these structures. We have demonstrated here and in ongoing work that procedural knowledge encoded across both the knowledge graph and reasoning codebase is a viable approach to building information structures. The real challenge lies in how we give AIDA the ability to \textit{recognize} when to use such structures and what elements should be involved. We are working on this question now.

A second important question that we are actively pursuing is how we can cast procedural knowledge into only a set of declarative statements, if it is possible at all. Doing so would allow AIDA to have all knowledge about not only how to write software, but how to write her own software, including her reasoning codebase, from scratch upon startup using only a very simple bootstrap program. In principle, SPARQL queries, which are the heart of the logic in the reasoning codebase, have structure and purpose very much like software generally. Thus, they should be amenable to reduction to a representation that can be encoded into a knowledge graph. Writing her own knowledge graph queries from scratch as needed itself would be another significant accomplishment for AIDA.

% Use \bibliography{yourbibfile} instead or the References section will not appear in your paper
\bibliography{matties-aida-arxiv}

\begin{thebibliography}{10}
\providecommand{\natexlab}[1]{#1}

\bibitem[{Ehrlinger and W{\"o}{\ss}(2016)}]{ehrlinger2016towards}
Ehrlinger, L.; and W{\"o}{\ss}, W. 2016.
\newblock Towards a Definition of Knowledge Graphs.
\newblock In \emph{SEMANTiCS (Posters, Demos, SuCCESS)}.

\bibitem[{{{GitHub Copilot}}(2022)}]{copilot:1}
{{GitHub Copilot}}. 2022.
\newblock GitHub Copilot.
\newblock \url{https://github.com/features/copilot}.
\newblock Accessed: 2022-06-30.

\bibitem[{{{GitHub Copilot FAQ}}(2022)}]{copilot:2}
{{GitHub Copilot FAQ}}. 2022.
\newblock GitHub Copilot ``Frequently asked questions'' under the question
  ``Does GitHub Copilot write perfect code?''.
\newblock \url{https://github.com/features/copilot}.
\newblock Accessed: 2022-06-30.

\bibitem[{{{Kite}}(2022)}]{kite:1}
{{Kite}}. 2022.
\newblock Kite.
\newblock \url{https://www.kite.com}.
\newblock Accessed: 2022-06-30.

\bibitem[{Krötzsch, Simancik, and Horrocks(2012)}]{abs-1201-4089}
Krötzsch, M.; Simancik, F.; and Horrocks, I. 2012.
\newblock A Description Logic Primer.
\newblock \emph{CoRR}, abs/1201.4089.

\bibitem[{Musen(2015)}]{protege-Musen15}
Musen, M.~A. 2015.
\newblock The protégé project: a look back and a look forward.
\newblock \emph{AI Matters}, 1(4): 4--12.

\bibitem[{{{OpenAI Codex}}(2022)}]{codex:1}
{{OpenAI Codex}}. 2022.
\newblock OpenAI Codex.
\newblock \url{https://openai.com/blog/openai-codex}.
\newblock Accessed: 2022-06-30.

\bibitem[{Pavese(2022)}]{sep-knowledge-how}
Pavese, C. 2022.
\newblock {Knowledge How}.
\newblock In Zalta, E.~N., ed., \emph{The {Stanford} Encyclopedia of
  Philosophy}. Metaphysics Research Lab, Stanford University, {F}all 2022
  edition.

\bibitem[{Singhal(2012)}]{singhal2012introducing}
Singhal, A. 2012.
\newblock Introducing the Knowledge Graph: things, not strings.
\newblock 2020-11-13.

\bibitem[{W3C(2012)}]{OWL2Profiles}
W3C. 2012.
\newblock OWL 2 Web Ontology Language Profiles (Second Edition).
\newblock Http://www.w3.org/TR/owl2-profiles/.

\end{thebibliography}

\section{Acknowledgments}
%% comment out the full acknowledgment and uncomment the line below before submitting for review!!
%%\textit{Redacted for review.}
The author thanks Dr. Amanda Hicks of JHU APL for many valuable discussions and Mr. Mark LoPresto, Mr. Andrew Adams, Dr. Scott Laprise, and Dr. Andrew Newman, all of JHU APL, for their valuable feedback in review of this work.

\end{document}